\documentclass[usletter, 10pt, conference]{ieeeconf}

\usepackage{url}

\usepackage[ruled,noend,linesnumbered]{algorithm2e}
\usepackage{graphicx,color,psfrag}
\usepackage{amsmath}
\usepackage{amsfonts}
\usepackage{amssymb}
\usepackage{bigdelim}
\usepackage{dsfont}
\usepackage{color, soul}
\usepackage{graphicx}
\usepackage{caption}
\usepackage{subcaption}

\IEEEoverridecommandlockouts              
\title{6D Object Pose Estimation from Accurate 3D Instance-aware Semantic Reconstructions for Warehouse Robots} 

\author
{Dinh-Cuong Hoang$^{*}$, Todor Stoyanov$^{*}$, and Achim J.   Lilienthal$^{*}$
\thanks{*Centre for Applied Autonomous Sensor Systems (AASS); Orebro University. Email: hoangcuongbk80@gmail.com
}
}

\begin{document}
\maketitle
\thispagestyle{empty}
\pagestyle{empty}


\begin{abstract}

We present an approach for recognizing all objects in a scene and estimating their full pose from an accurate 3D instance-aware semantic reconstruction using an RGB-D camera. Our framework couples convolutional neural networks (CNNs) and a state-of-the-art dense Simultaneous Localisation and Mapping (SLAM) system, ElasticFusion, to achieve both high-quality semantic reconstruction as well as robust 6D pose estimation for relevant objects. While the main trend in CNN-based 6D pose estimation has been to infer object's position and orientation from single views of the scene, our approach explores performing pose estimation from multiple viewpoints, under the conjecture that combining multiple predictions can improve the robustness of an object detection system. The resulting system is capable of producing high-quality object-aware semantic reconstructions of room-sized environments, as well as accurately detecting objects and their 6D poses. The developed method has been verified through experimental validation on the YCB-Video dataset and a newly collected warehouse object dataset. Experimental results confirmed that the proposed system achieves improvements over state-of-the-art methods in terms of surface reconstruction and object pose prediction. Our code and video are available at \textit{https://sites.google.com/view/object-rpe}.
\end{abstract}


\section{Introduction}
\label{sec:intro}

Simultaneous localization and mapping (SLAM) is a crucial enabling technology for autonomous warehouse robots. With the increasing availability of RGB-D sensors, research on visual SLAM has made giant strides in development \cite{newcombe2011kinectfusion, kerl2013robust, whelan2016elasticfusion}. These approaches achieve dense surface reconstruction of complex and arbitrary indoor scenes while maintaining real-time performance through implementations on highly parallelized hardware. However, the purely geometric map of the environment produced by classical SLAM systems is not sufficient to enable robots to operate safely and effectively in warehouse applications with a high demand on flexibility. For instance, automated picking and manipulation of boxes and other types of goods requires information about the position and orientation of objects.  The inclusion of rich semantic information and 6D poses of object instances within a dense map is required to help robots better understand their surroundings, to avoid undesirable contacts with the environment and to accurately grasp selected objects.
\par
Beyond classical SLAM systems which solely provide a purely geometric map, the idea of a system that generates a dense map in which object instances are semantically annotated has attracted substantial interest in the research community \cite{hoang2019high,
sunderhauf2017meaningful, mccormac2018fusion++, runz2018maskfusion}. Semantic 3D maps are important for robotic scene understanding, planning and interaction. In the case of automated warehouse picking, providing accurate object poses together with semantic information are crucial for robots that have to manipulate the objects around them in diverse ways.
\par
To obtain the 6D pose of objects, many approaches were introduced in the past \cite{fuchs2010cooperative, corney2002coarse, germann2007automatic}. However, because of the complexity of object shapes, measurement noise and presence of occlusions, these approaches are not robust enough in real applications. Recent work has attempted to leverage the power of deep CNNs to solve this nontrivial problem \cite{xiang2017posecnn, tekin2018real, wang2019densefusion}. These techniques demonstrate a significant improvement of the accuracy of 6D object pose estimation on some popular datasets such as YCB-Video or LineMOD. Even so, due to the limitation of single-view-based pose estimation, the existing solutions generally do not perform well in cluttered environments and under large occlusions. 
\par
In this work, we develop a system for 6D objects pose estimation that benefits from the use of our accurate intance-aware semantic mapping system and from combining multiple predictions. Intuitively, by combining pose predictions from multiple camera views, the accuracy of the estimated 3D object pose can be improved. Based on this, our framework deploys simultaneously a 3D mapping algorithm to reconstruct a semantic model of the environment, and an incremental 6D object pose recovering algorithm that carries out predictions using the reconstructed model. We demonstrate that we can exploit multiple viewpoints around the same object to achieve robust and stable 6D pose estimation in the presence of heavy clutter and occlusion.
\par
Our main contribution is, therefore, a method that can be used to accurately predict the pose of objects under partial occlusion. We demonstrate that by integrating deep learning-based pose prediction into our semantic mapping system we are able to address the challenges posed by missing information due to clutter, self-occlusions, and bad reflections. 
\section{RELATED WORK}

In recent years, CNN architectures have been extended to the object pose estimation task \cite{xiang2017posecnn, tekin2018real, wang2019densefusion}. SingleShotPose \cite{tekin2018real} simultaneously detects an object in an RGB image and predicts its 6D pose without requiring multiple stages or having to examine multiple hypotheses. It is end-to-end trainable and only needs the 3D bounding box of the object shape for training. This method is able to deal with textureless objects, however, it fails to estimate object poses under large occlusions. To handle occlusions better, the PoseCNN architecture \cite{xiang2017posecnn} employs semantic labeling which provides richer information about the objects. PoseCNN recovers the 3D translation of an object by localizing its center in the image and estimating the 3D center distance from the camera. The 3D rotation of the object is estimated by regressing convolutional features to a quaternion representation. In addition, in order to handle symmetric objects, the authors introduce ShapeMatch-Loss, a new loss function that focuses on matching the 3D shape of an object. The results show that this loss function produces superior estimation for objects with shape symmetries. However, this approach requires Iterative Closest Point (ICP) for refinement which is prohibitively slow for real-time applications. To solve this problem, Wang et al. proposed DenseFusion \cite{wang2019densefusion} which is approximately 200x faster than PoseCNN-ICP and outperforms previous approaches in two datasets, YCB-Video and LineMOD. The key technique of DenseFusion is that it extracts features from the color and depth images and fuses RGB values and point clouds at the per-pixel level. This per-pixel fusion scheme enables the model to explicitly reason about the local appearance and geometry information, which is essential to handle occlusions between objects. In addition, an end-to-end iterative pose refinement procedure is proposed to further improve pose estimation while achieving near real-time inference. Although DenseFusion has achieved impressive results, like other single-view-based methods it suffers significantly from the ambiguity of object appearance and occlusions in cluttered scenes, which are very common in practice. In addition, since DenseFusion relies on segmentation results for pose prediction, its accuracy highly depends on the performance of the segmentation framework used. As in pose estimation networks, if the input to a segmentation network contains an occluder, the occlusion significantly influences the network output. In this paper, while exploiting the advantages of the DenseFusion framework, we replace its segmentation network by our semantic mapping system that provides a high-quality segmentation mask for each instance. We address the problem of the ambiguity of object appearance and occlusion by combining predictions using RGB-D images from multiple viewpoints.
%

\section{METHODOLOGY}
\label{sec:methodology}

\begin{figure}
\centering
	\includegraphics[width = 0.85\linewidth, height=4.0cm]{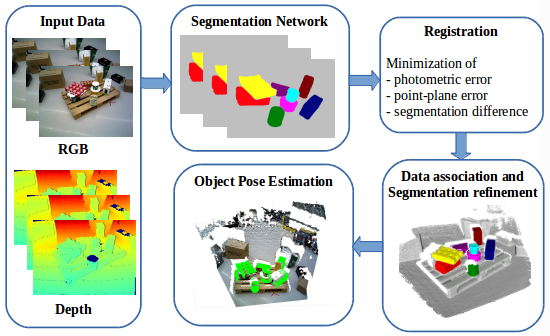}
\caption{Overview of the proposed system.}
\label{fig:overview}
\end{figure}

Our pipeline is composed of four main components as illustrated in Fig.~\ref{fig:overview}. Input data is processed through a segmentation network followed by a registration stage. Using the estimated sensor pose, the dense 3D geometry of the map or model is updated by fusing the points labeled in the fusion stage. The last component is 6D object pose estimator that output the pose of objects by combining predictions from single-view-based predictions. In the following, we summarise the key elements of our method.

\textbf{Segmentation}: The network takes in RGB images (only keyframes) and extracts instance masks labeled with object class, which serve as input to the subsequent registration and fusion stages.

\textbf{Registration}: Estimate camera poses within the ElasticFusion pipeline using a joint cost function that combine the cost functions of geometric and photometric estimates in a weighted sum.

\textbf{Data Fusion}: Our map representation is an unordered list of surfels similar to \cite{whelan2016elasticfusion}. The surfel map is updated by merging the newly available RGB-D frame into the existing models. In addition, segmentation information is fused into the map using our instance-based semantic fusion scheme. To improve segmentation accuracy, misclassified regions are corrected by two criteria which rely on a sequence of CNN
predictions.

\textbf{Object Pose Estimation}: First, we employ DenseFusion
that operates on object instances from single views to predict object poses. Instead of using depth and color frames captured by the camera, we use the surfel-splatted predicted depth map and the color image of the model from the previous pose estimate for DenseFusion. The predicted poses are then used as a measurement update in a Kalman filter to achieve optimal 6D pose of objects.

\begin{figure*}[h!]
  \centering
  
  \begin{subfigure}[b]{0.2\linewidth}
    \includegraphics[width=3.2cm, height=2.3cm]{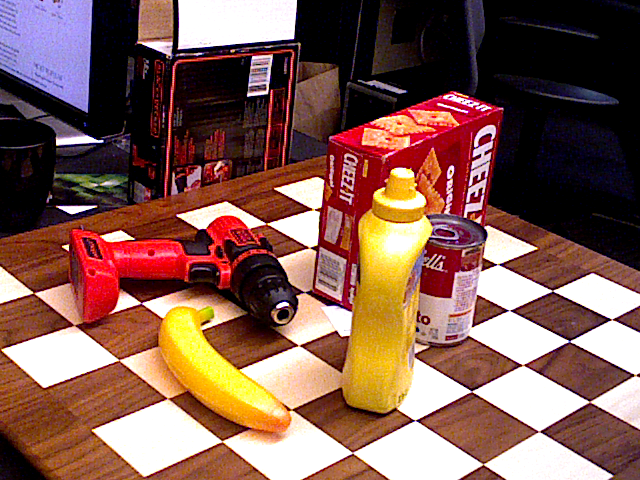}
  \caption{frame 66}  
  \end{subfigure}
  \begin{subfigure}[b]{0.2\linewidth}
    \includegraphics[width=3.2cm, height=2.3cm]{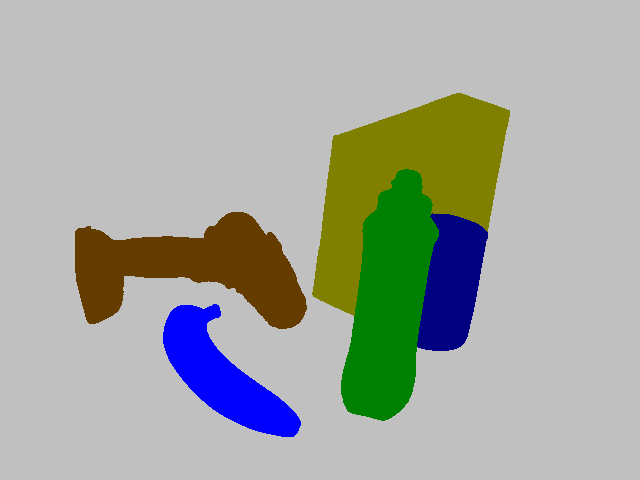}
  \caption{Ground truth}  
  \end{subfigure}  
  \begin{subfigure}[b]{0.2\linewidth}
    \includegraphics[width=3.2cm, height=2.3cm]{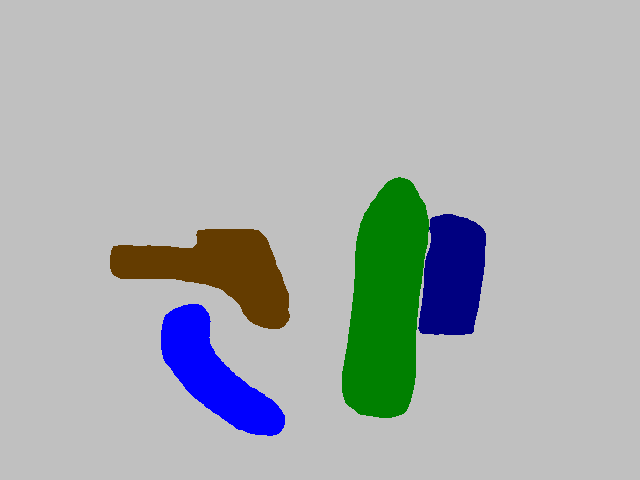}
  \caption{Mask R-CNN}  
  \end{subfigure}  
  \begin{subfigure}[b]{0.2\linewidth}
    \includegraphics[width=3.2cm, height=2.3cm]{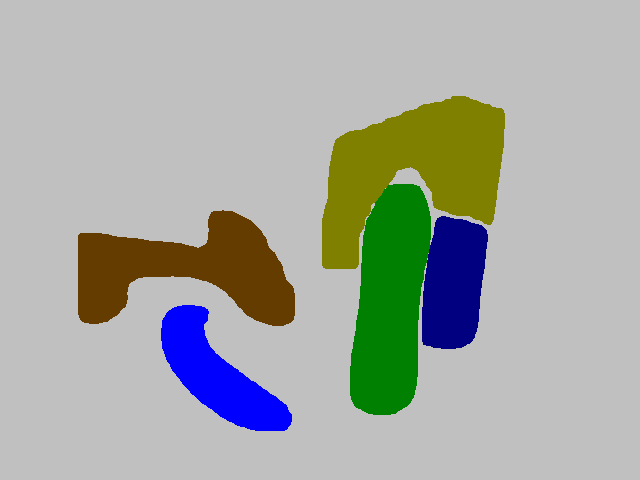}
  \caption{Object-RPE}  
  \end{subfigure} 
  
  \begin{subfigure}[b]{0.2\linewidth}
    \includegraphics[width=3.2cm, height=2.3cm]{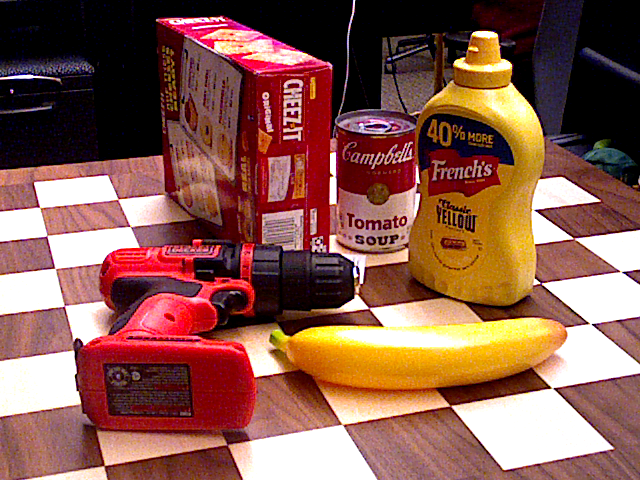}
  \caption{frame 1916}  
  \end{subfigure} 
  \begin{subfigure}[b]{0.2\linewidth}
    \includegraphics[width=3.2cm, height=2.3cm]{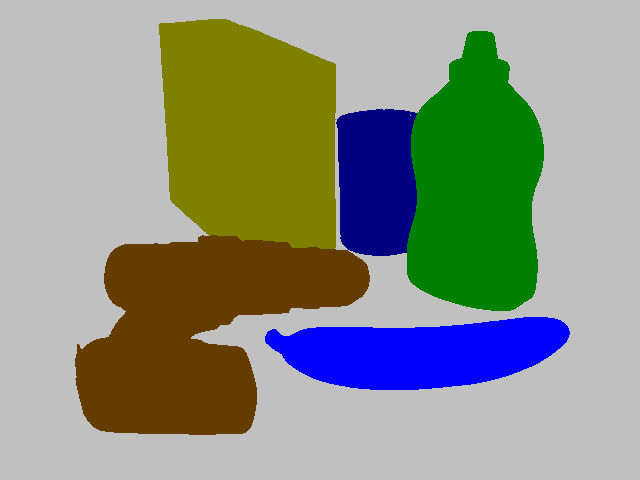}
  \caption{Ground truth}  
  \end{subfigure}  
  \begin{subfigure}[b]{0.2\linewidth}
    \includegraphics[width=3.2cm, height=2.3cm]{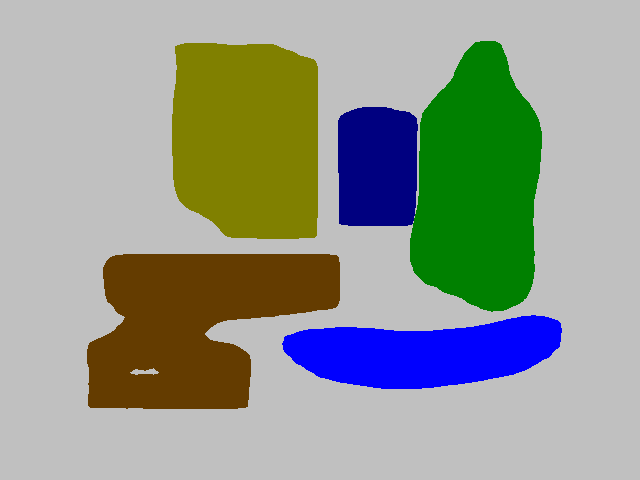}
  \caption{Mask R-CNN}  
  \end{subfigure}          
  \begin{subfigure}[b]{0.2\linewidth}
    \includegraphics[width=3.2cm, height=2.3cm]{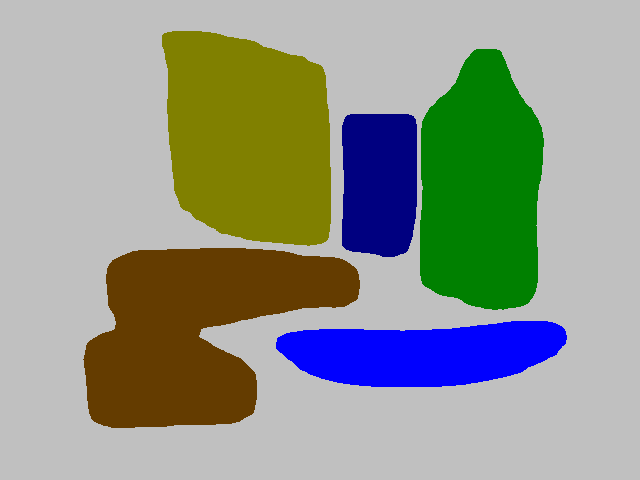}
  \caption{Object-RPE}  
  \end{subfigure}
  
  \caption{Examples of masks generated by Mask R-CNN and produced by reprojecting the current scene model.}
  \label{fig:segmentation_results}
\end{figure*}


\subsection{Instace-aware Semantic Mapping}
\label{sec:mapping}

\textbf{Segmentation}: We employ an end-to-end CNN framework, Mask R-CNN \cite{he2017mask} for generating a high-quality segmentation mask for each instance. Mask R-CNN has three outputs for each candidate object, a class label, a bounding box offset, and a mask. Its procedure consists of two stages. In the first stage, candidate object bounding boxes are proposed by a Region Proposal Network (RPN). In the second stage, classification, bounding-boxregression, and mask prediction are performed in parallel oneach small feature map, which is extracted by RoIPool. Note that to speed up inference and improve accuracy the mask branch is applied to the highest scoring 100 detection boxes after running the box prediction. The mask branch predicts a binary mask from each RoI using an FCN architecture \cite{long2015fully}. The binary mask is a single $ m \times m $ output regardless of class, which is generated by binarizing the floating-number mask or soft mask at a threshold of 0.5.

\textbf{Registration:} Similar to ElasticFusion, our approach aims to estimate a sensor pose that minimizes the cost over a combination of the global point-plane energy and photometric error. We wish to minimize a joint optimization objective:

\begin{align}
E_{combined} = E_{icp} + \omega E_{rgb} \
\label{eq:cost_function}
\end{align}

\noindent where $E_{icp}$ and $\omega E_{rgb}$ are the geometric and photometric error terms respectively.

\textbf{Data association:} Given an RGB-D frame at time step $t$, each mask $M$ from Mask R-CNN must be corresponded to an instance in the 3D map. Otherwise, it will be assigned as a new instance. To find the corresponding instance, we use the tracked camera pose and existing instances in the map built at time step $t-1$ to predict binary masks via splatted rendering. The percent overlap between the mask $M$ and a predicted mask $\hat{M}$ for object instance $\mathbf{o}$ is computed as $\mathbb{U}(M, \hat{M})=\dfrac{M\cap\hat{M}}{\hat{M}}$. Then the mask $M$ is mapped to object instance $\mathbf{o}$ which has the predicted mask $\hat{M}$ with largest overlap, where $\mathbb{U}(M, \hat{M}) > 0.3$.

To efficiently store class probabilities, we propose to assign an object instance label $\mathbf{o}$ to each surfel and then this label is associated with a discrete probability distribution over potential class labels, $P(L_{\mathbf{o}}=l_{i})$ over the set of class labels, $l_{i}\in\mathbb{L}$. In consequence, we need only one probability vector for all surfels belonging to the same object entity. This makes a big difference when the number of surfels is much larger than the number of classes. To update the class probability distribution, means of a recursive Bayesian update is used in \cite{hermans2014dense}. However, this scheme often results in an overly confident class probability distribution that contains scores unsuitable for ranking in object detection \cite{mccormac2018fusion++}. In order to make the distribution become more even, we update the class probability by simple averaging:

\begin{align}
\ P(l_{i}|I_{1,..,t}) = \dfrac{1}{t} \sum_{j=1}^{t}(p_{j}|I_{t}) \
\label{eq:fusion_eq}
\end{align}

Moreover, previous related works miss the background/object probability from the binary mask branch that predicts which pixels correspond to the main classes (non-background), and which pixels correspond to the background. Conversely, we enrich segmentation information on each surfel by adding the probability to account for background/object predictions. To that end, each surfel in our 3D map has a non-background probability attribute $ p_{o} $.

As presented in \cite{he2017mask} the binary mask branch first generates a $m \times m $ floating-number mask which is then resized to the RoI size, and binarized at a threshold of 0.5. Therefore, we are able to extract a per-pixel non-background probability map with the same image size $480 \times 640$. Given the RGB-D frame at time step $t$, a non-background probability $ p_{\mathbf{o}}(I_{t}) $ is assigned to each pixel. Camera tracking and the 3D back projection introduced in section enables us to update all the surfels with the corresponding probability as following:

\begin{align}
\ p_{\mathbf{o}} = \frac{1}{t} \sum_{j=1}^{t}p_{j}(I_{t})\
\label{eq:fusion_eq}
\end{align}

\textbf{Segmentation Improvement:} Despite the power and flexibility of Mask R-CNN, it usually misclassified object boundary regions as background. In other words, the detailed structures of an object are often lost or smoothed. Thus, there is still much room for improvement in segmentation. We observe that many of the pixels in the misclassified regions have non-background probability just slightly smaller than 0.5, while the soft probabilities mask for real background pixel is often far below the threshold. Based on this observation, we expect to achieve a more accurate object-aware semantic scene reconstruction by considering non-background probability of surfels within a $n$ frames sequence. With this goal, each possible surfel $s$ ($0.4 < p_{\mathbf{o}} < 0.5$) is associated with a confidence $\vartheta(s)$. If a surfel is identified for the first time, its associated confidence is initialized to zero. Then, when a new frame arrives, we increment the confidence $\vartheta(s) \leftarrow \vartheta(s)+1$ only if the corresponding pixel of that surfel satisfies 2 criteria: (i) its non-background probability is greater than 0.4; (ii) there is at least one object pixel inside its 6-neighborhood. After $n$ frames, if the confidence $\vartheta(s)$ exceeds the threshold $\sigma_{object}$, we assign surfel $s$ to the closest instance. Otherwise, $\vartheta(s)$ is reset to zero. Here, we found $n=10$ and $\sigma_{object}=10$ provide good performance.

\subsection{Multi-view Object Pose Estimation}
\label{sec:Multi_Pose_Estimation}

Given an RGB-D frame sequence, the task of 6D object pose estimation is to estimate the rigid transformation from the object coordinate system $\mathcal{O}$ to a global coordinate system $\mathcal{G}$. We assume that the 3D model of the object is available and the object coordinate system is defined in the 3D space of the model. The rigid transformation consists of a 3D rotation $R(\omega, \varphi, \psi)$ and a 3D translation $T(X, Y, Z)$. The translation $T$ is the coordinate of the origin of $\mathcal{O}$ in the global coordinate frame $\mathcal{G}$, and $R$ specifies the rotation angles around the X-axis, Y-axis, and Z-axis of the object coordinate system $\mathcal{O}$.
\par
Our approach outputs the object poses with respect to the global coordinate system by combining predictions from different viewpoints. For each frame at time $t$, we apply DenseFusion to masks back-projected from the current 3D map. The estimated object poses are then transferred to the global coordinate system $\mathcal{G}$ and serve as measurement inputs for an extended Kalman filter (EKF) based pose update stage. 

\textbf{Single-view based prediction:} In order to estimate the pose of each object in the scene from single views, we apply DenseFusion to masks back-projected from the current 3D map. The network architecture and hyperparameters are similar as introduced in the original paper \cite{wang2019densefusion}. The image embedding network consists of a ResNet-18 encoder followed by 4 up-sampling layers as a decoder. The PointNet architecture is a multi-layer perceptron (MLP) followed by an average-pooling reduction function. The iterative pose refinement module consists of 4 fully connected layers that directly output the pose residual from the global dense feature. For each object instance mask, a 3D point cloud is computed from the predicted model depth pixels and an RGB image region is cropped by the bounding box of the mask from the predicted model color image. First, the image crop is fed into a fully convolutional network and then each pixel is mapped to a color feature embedding. For the point cloud, a PointNet-like architecture is utilized to extract geometric features. Having generated features, the next step combines both embeddings and outputs the estimation of the 6D pose of the object using a pixel-wise fusion network. Finally, the pose estimation results are improved by a neural network-based iterative refinement module. A key distinction between our approach and DenseFusion is that instead of directly operating on masks from the segmentation network, we use predicted 2D masks that are obtained by reprojecting the current scene model. As illustrated in Fig.~\ref{fig:segmentation_results} our semantic mapping system leads to an improvement in the 2D instance labeling over the baseline single frame predictions generated by Mask R-CNN. As a result, we expect that our object pose estimation method benefits from the use of the more accurate segmentation results.

\textbf{Object pose update:} For each frame at time $t$, the estimates obtained by DenseFusion and camera motions from the registration stage are used to compute the pose of each object instance  with respect to the global coordinate system $\mathcal{G}$. The pose is then used as a measurement update in a Kalman filter to estimate an optimal 6D pose of the object. Since we assume that the measured scene is static over the reconstruction period, the object's motion model is constant. The state vector of the EKF combines the estimates of translation and rotation:
\begin{align}
\ \textbf{x} = [X\quad Y\quad Z\quad \phi\quad \varphi\quad \psi]^{\top} \
\label{eq:state_vec}
\end{align}

Let $x_{t}$ be the state at time $t$, $\hat{\textbf{x}}^{-}_{t}$ denote the predicted state estimate and $P^{-}_{t}$ denote predicted error covariance at time $t$ given the knowledge of the process and measurement at the end of step $t-1$, and let $\hat{\textbf{x}}_{t}$ be the updated state estimate at time $t$ given the pose estimated by DenseFusion $z_{t}$. The EKF consists of two stages prediction and measurement update (correction) as follows.

Prediction:
\begin{align}
\ \hat{\textbf{x}}^{-}_{t} = \hat{\textbf{x}}_{t-1} \\
\ P^{-}_{t} = P_{t-1}
\end{align}

Measurement update:
\begin{align}
\hat{\textbf{x}}_{t} & = \hat{\textbf{x}}^{-}_{t} \oplus K_{t}(z_{t} \ominus \hat{\textbf{x}}^{-}_{t}) \\
K_{t} & = P^{-}_{t} (R_{t} + P^{-}_{t})^{-1} \\
P_{t} & = (I_{6\times6} - K_{t}) P^{-}_{t}
\end{align}

Here, $\ominus$ and $\oplus$ are the pose composition operators. $K_{t}$ is the Kalman gain update. The $6\times6$ matrix $R_{t}$ is measurement noise covariance, computed as:
\begin{align}
\ R_{t} = \mu I_{6\times6}
\end{align}

\noindent where $\mu$ is the average distance of all segmented object points from the corresponding 3D model points transformed according to the estimated pose.
\section{EXPERIMENTS}
\label{sec:experiments}

We evaluated our system on the YCB-Video \cite{xiang2017posecnn} dataset and on a newly collected warehouse object dataset. The YCB-Video dataset was split into 80 videos for training and the remaining 12 videos for testing. For the warehouse object dataset, the system was trained on 15 videos and tested on the other 5 videos. Our experiments are aimed at evaluating both surface reconstruction and 6D object pose estimation accuracy. A comparison against the most closely related works is also performed here.

For all tests, we ran our system on a standard desktop PC running 64-bit Ubuntu 16.04 Linux with an Intel Core i7-4770K 3.5GHz and a nVidia GeForce GTX 1080 Ti 6GB GPU. Our pipeline is implemented in C++ with CUDA for RGB-D image registration. The Mask R-CNN and DenseFusion codes are based on the publicly available implementations by Matterport\footnote{\url{https://github.com/matterport/Mask_RCNN}} and Wang\footnote{\url{https://github.com/j96w/DenseFusion}}. In all of the presented experimental setups, results are generated from RGB-D video with a resolution of 640x480 pixels. The DenseFusion networks were trained for 200 epochs with a batchsize of 8. Adam \cite{kingma2014adam} was used as the optimizer with learning rate set to 0.0001.

\begin{figure}[thb]
  \centering
  
  \begin{subfigure}[b]{0.24\linewidth}
    \includegraphics[width=1.8cm]{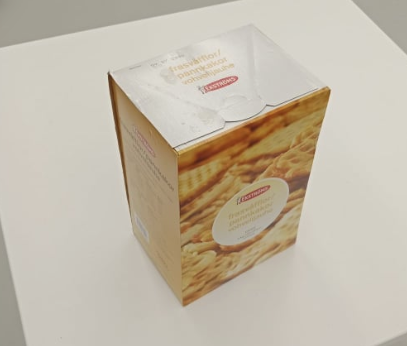}
    \caption{Waffle}
  \end{subfigure}
  \begin{subfigure}[b]{0.24\linewidth}
    \includegraphics[width=1.8cm]{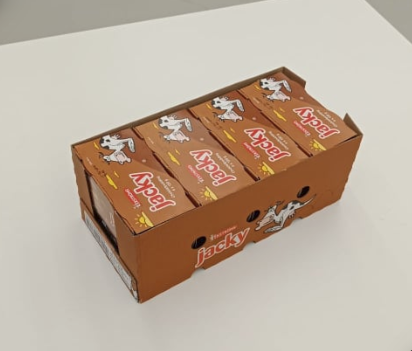}
    \caption{Jacky}
  \end{subfigure}
  \begin{subfigure}[b]{0.24\linewidth}
    \includegraphics[width=1.8cm]{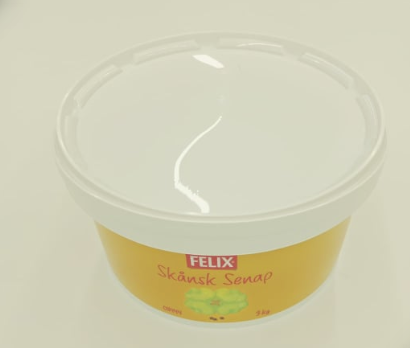}
    \caption{Skansk}
  \end{subfigure}
  \begin{subfigure}[b]{0.24\linewidth}
    \includegraphics[width=1.8cm]{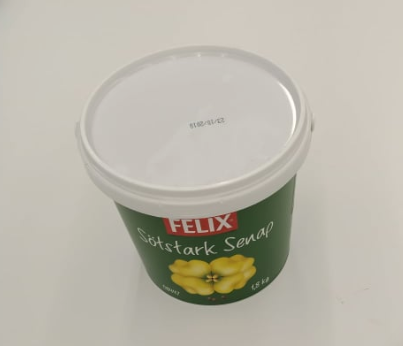}
    \caption{Sotstark}
  \end{subfigure}
  \begin{subfigure}[b]{0.24\linewidth}    
    \includegraphics[width=1.8cm]{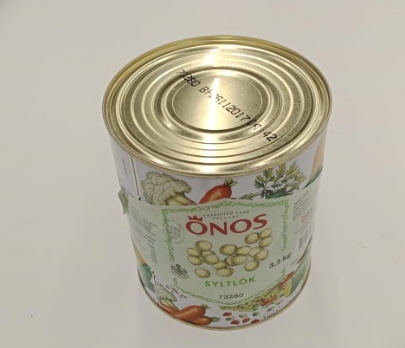}
    \caption{Onos}
  \end{subfigure}
  \begin{subfigure}[b]{0.24\linewidth}    
    \includegraphics[width=1.8cm]{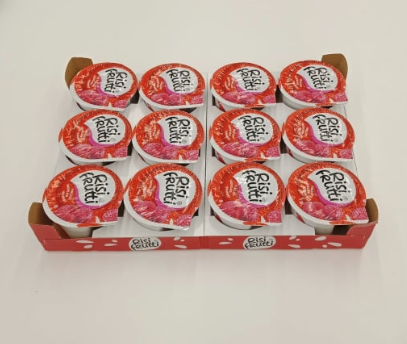}
    \caption{Risi Frutti}
  \end{subfigure}
  \begin{subfigure}[b]{0.24\linewidth}
    \includegraphics[width=1.8cm]{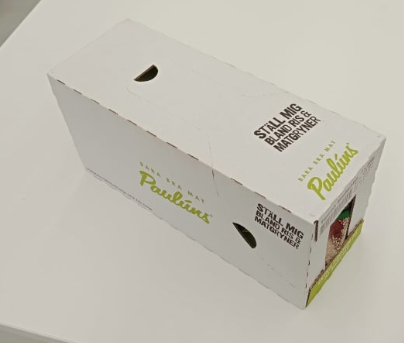}
    \caption{Pauluns}
  \end{subfigure}    
  \begin{subfigure}[b]{0.24\linewidth}
    \includegraphics[width=1.8cm]{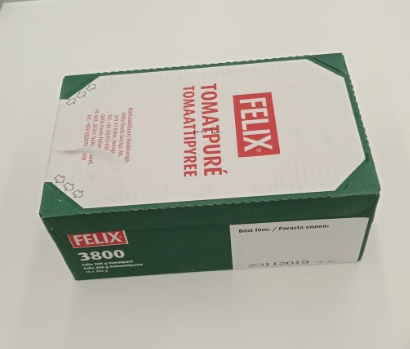}
    \caption{Tomatpure}
  \end{subfigure}
  \begin{subfigure}[b]{0.28\linewidth}
    \includegraphics[width=2.2cm]{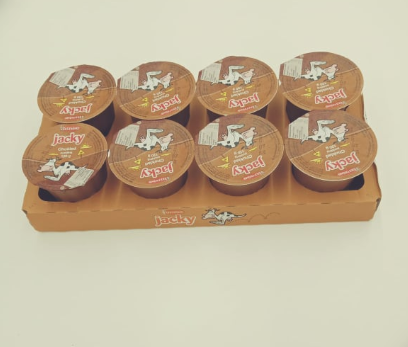}
    \caption{Small Jacky}
  \end{subfigure}  
  \begin{subfigure}[b]{0.28\linewidth}
    \includegraphics[width=2.2cm]{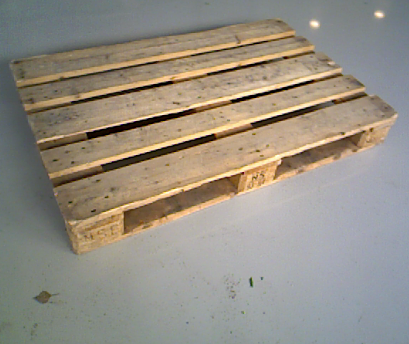}
    \caption{Pallet}
  \end{subfigure}
  \begin{subfigure}[b]{0.28\linewidth}
    \includegraphics[width=2.2cm]{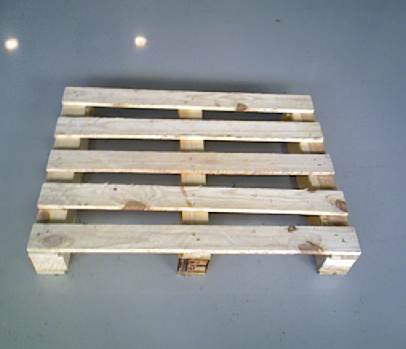}
    \caption{Half Pallet}
  \end{subfigure}
  
  \caption{The set of 11 objects in the warehouse object dataset.}
  \label{fig:warehouse_objects}
\end{figure}


\begin{figure}[t!]
  \centering
  
  \begin{subfigure}[b]{0.33\linewidth}
    \includegraphics[width=2.5cm]{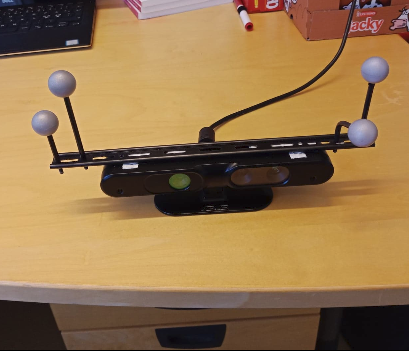}
    \caption{}
  \end{subfigure}
  \begin{subfigure}[b]{0.32\linewidth}
    \includegraphics[width=2.5cm]{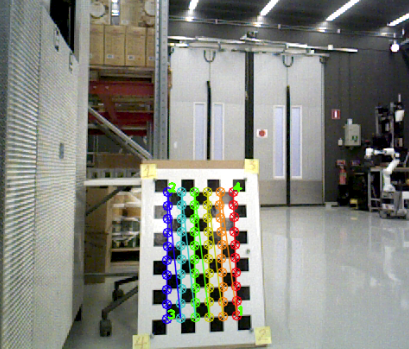}
    \caption{}
  \end{subfigure}
  \begin{subfigure}[b]{0.32\linewidth}
    \includegraphics[width=2.5cm]{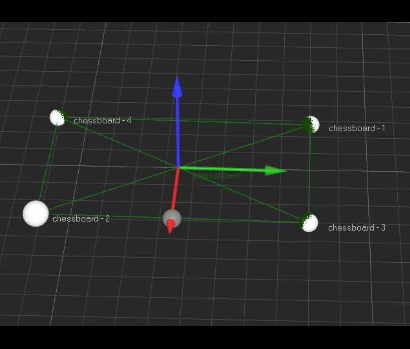}
    \caption{}
  \end{subfigure}
  
  \caption{We collected a dataset for the evaluation of reconstruction and pose estimation systems in a typical warehouse using (a) a hand-held ASUS Xtion PRO LIVE sensor. Calibration parameters were found by using (b) a chessboard and (c) reflective markers detected by the motion capture system.}
  \label{fig:dataset_collecting}
\end{figure}

\subsection{The Warehouse Object Dataset}
\label{dataset}
Unlike scenes recorded in the YCB-Video dataset or other publicly
available datasets, warehouse environments pose more complex problems, including low illumination inside shelves, low-texture and symmetric objects, clutter, and occlusions. To advance warehouse application of robotics as well as to thoroughly evaluate our method, we collected an RGB-D video dataset of 11 objects as shown Fig.~\ref{fig:warehouse_objects}, which is focused on the challenges in detecting warehouse object poses using an RGB-D sensor. The dataset consists of over 20,000 RGB-D images extracted from 20 videos captured by an ASUS Xtion PRO Live sensor, the 6D poses of the objects and instance segmentation masks generated using the LabelFusion framework \cite{marion2018label}, as well as camera trajectories from a motion capture system developed by Qualisys\footnote{\url{https://www.qualisys.com}}. Calibration is required for both the RGB-D sensor and motion capture system shown in Fig.~\ref{fig:dataset_collecting}. We calibrated the motion capture system using the Qualisys Track Manager (QTM) software. For RGB-D camera calibration, the intrinsic camera parameters were estimated using classical black-white chessboard and the OpenCV library. In order to track the camera pose through the motion capture system, we attached four spherical markers on the sensor. In addtion, another four markers were also placed on the outer corners of a calibration checkerboard. By detecting these markers, we were able to estimate the transformation between the pose from the motion capture system and the optical frame of the RGB-D camera.

\begin{figure}[t!]
  \centering
  
  \begin{subfigure}[b]{0.45\linewidth}
    \includegraphics[width=3.5cm, height=3cm]{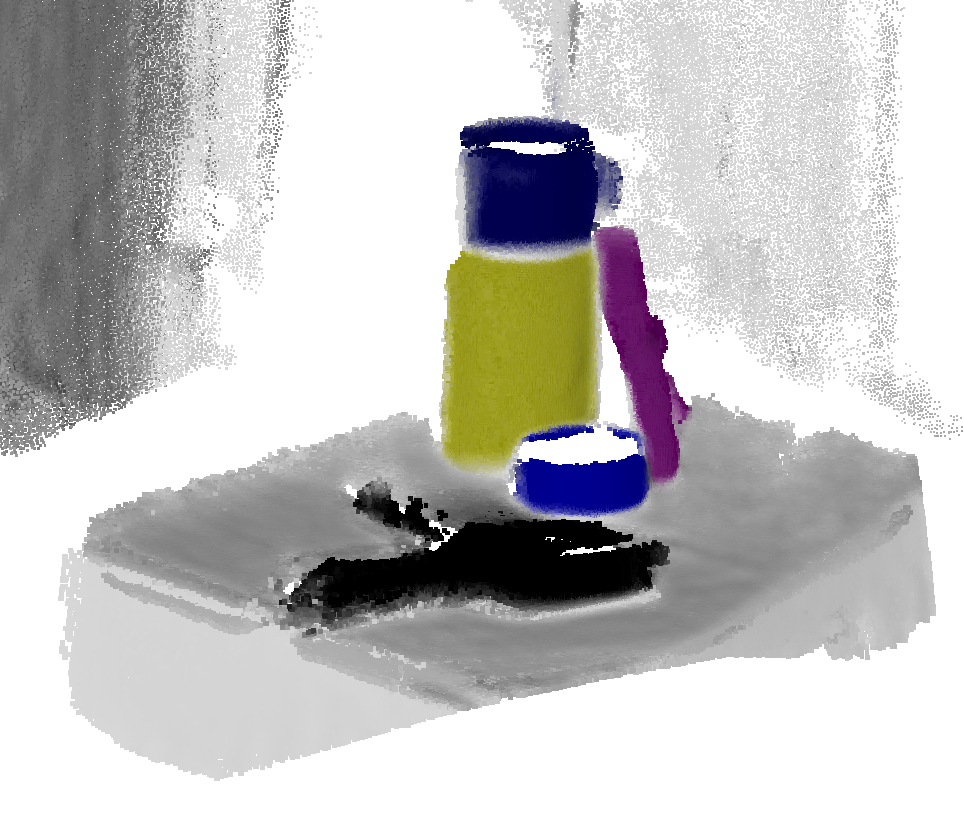}
  \end{subfigure}
  \begin{subfigure}[b]{0.45\linewidth}
    \includegraphics[width=3.5cm, height=3cm]{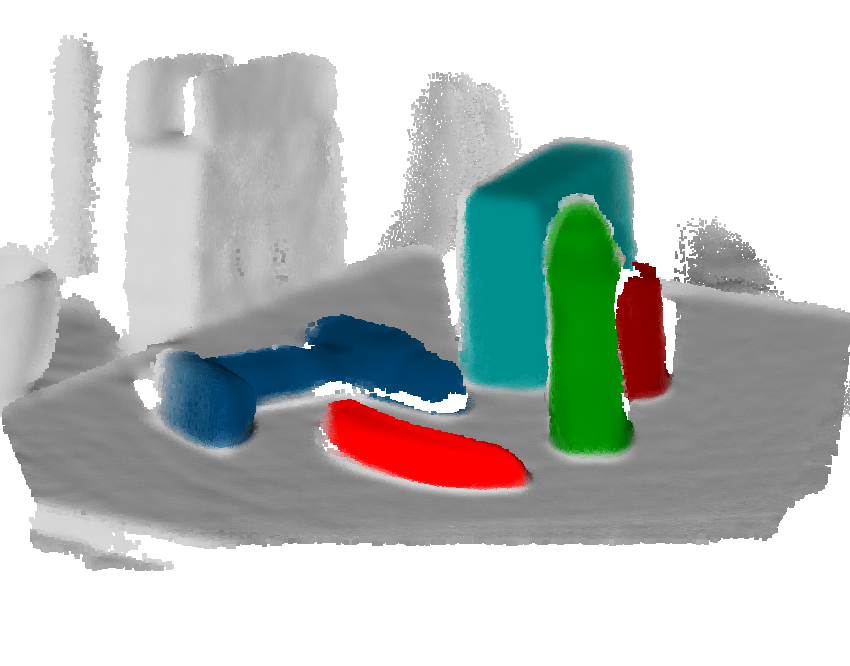}
  \end{subfigure}
  \begin{subfigure}[b]{0.45\linewidth}
    \includegraphics[width=3.5cm, height=3cm]{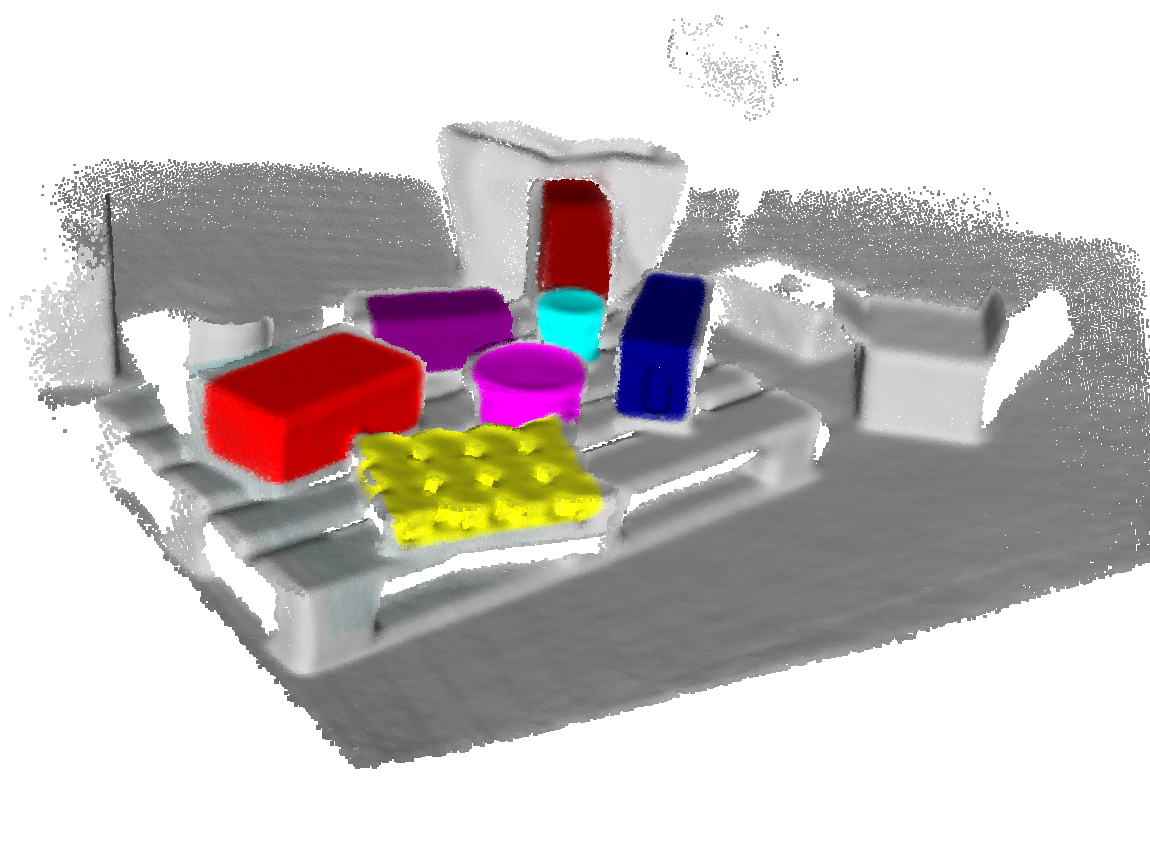}
  \end{subfigure}
  \begin{subfigure}[b]{0.45\linewidth}
    \includegraphics[width=3.5cm, height=3cm]{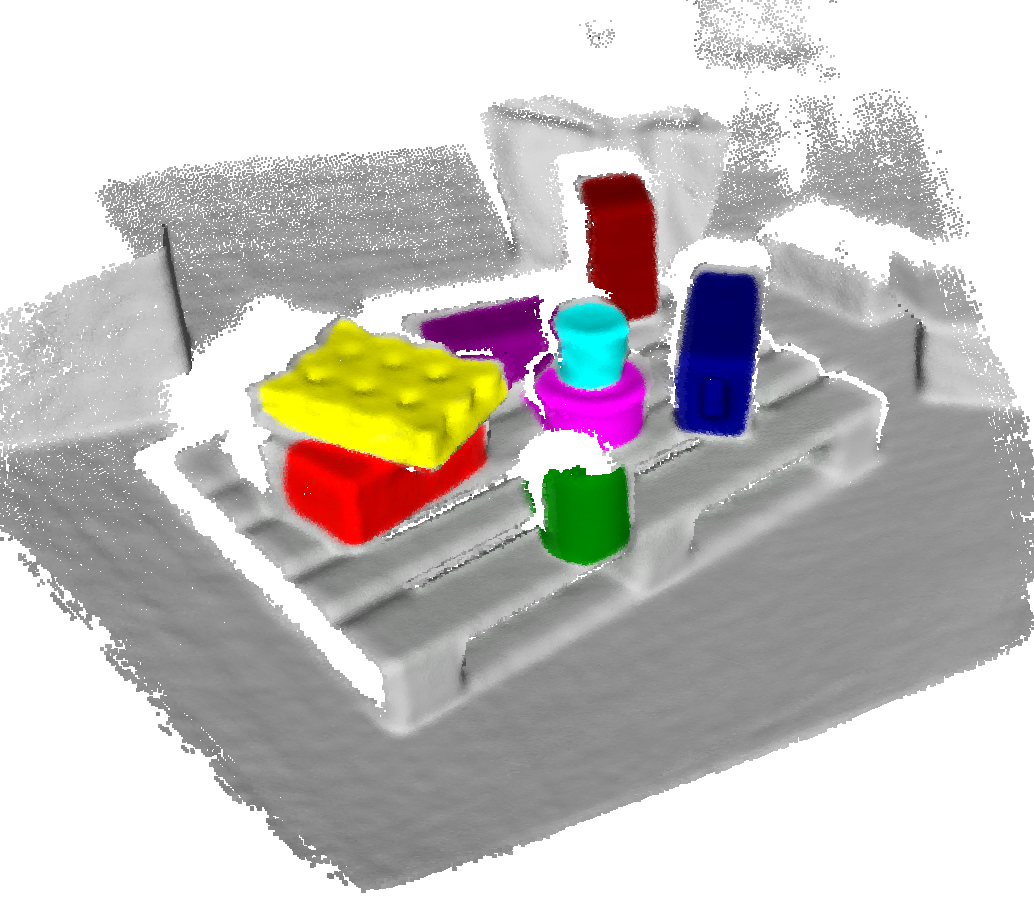}
  \end{subfigure}
  
  \caption{Examples of 3D object-aware semantic maps from the YCB-Video dataset and the warehouse object dataset.}
  \label{fig:semantic_maps}
\end{figure}

\subsection{Reconstruction Results}
\label{sec:reconstruc_result}

\begin{table*}[h]
\caption{Comparison of surface reconstruction error and pose estimation accuracy results on the YCB objects.}
\label{tab:results-ycb}
\begin{center}
\begin{tabular}{|l|c|c|c|c|c|c|c|}
\hline
& \multicolumn{2}{c|}{Reconstruction (mm)} & \multicolumn{5}{c|}{6D Pose Estimation} \\
\hline
& ElasticFusion & Object-RPE & DenseFusion (DF) & DF-PM & DF-PM-PD & DF-PM-PD-PC & Object-RPE \\
\hline
002\_master\_chef\_can & 5.7 & \textbf{4.5} & 96.4 & 96.8 & 96.5 & 97.0 & \textbf{97.6} \\
003\_cracker\_box & 5.2 & \textbf{4.8} & 95.5 & 96.2 & 96.2 & 96.9 & \textbf{97.3} \\
004\_sugar\_box & 7.2 & \textbf{5.3} & 97.5 & 97.4 & 97.0 & 97.2 & \textbf{98.1} \\
005\_tomato\_soup\_can & 6.4 & \textbf{5.7} & 94.6 & 94.7 & 95.2 & 95.6 & \textbf{96.8} \\
006\_mustard\_bottle & \textbf{5.2} & 6.1 & 97.2 & 97.7 & 97.9 & 97.9 & \textbf{98.3} \\
007\_tuna\_fish\_can & 6.8 & \textbf{5.4} & 96.6 & 97.1 & 97.4 & 98.1 & \textbf{98.5} \\
008\_pudding\_box & 5.6 & \textbf{4.3} & 96.5 & 97.3 & 97.1 & 97.6 & \textbf{98.4} \\
009\_gelatin\_box & 5.5 & \textbf{4.9} & 98.1 & 98.0 & 98.2 & 98.4 & \textbf{99.0} \\
010\_potted\_meat\_can & 7.4 & \textbf{6.3} & 91.3 & 92.2 & 92.5 & 92.9 & \textbf{94.7} \\
011\_banana & \textbf{6.2} & 6.4 & 96.6 & 96.8 & 96.8 & 97.4 & \textbf{97.9} \\
019\_pitcher\_base & 5.8 & \textbf{4.9} & 97.1 & 97.5 & 97.9 & 98.2 & \textbf{99.3} \\
021\_bleach\_cleanser & 5.4 & \textbf{4.2} & 95.8 & 96.5 & 95.9 & 96.3 & \textbf{97.6} \\
024\_bowl & 8.8 & \textbf{7.4} & 88.2 & 89.5 & 90.3 & 90.8 & \textbf{93.7} \\
025\_mug & 5.2 & \textbf{5.4} & 97.1 & 96.8 & 97.3 & 97.5 & \textbf{99.1} \\
035\_power\_drill & 5.8 & \textbf{5.1} & 96.0 & 96.6 & 96.8 & 96.8 & \textbf{98.1} \\
036\_wood\_block & 7.4 & \textbf{6.7} & 89.7 & 90.3 & 90.6 & 91.2 & \textbf{95.7} \\
037\_scissors & 5.5 & \textbf{5.1} & 95.2 & 96.2 & 96.2 & 96.2 & \textbf{97.9} \\
040\_large\_marker & 6.1 & \textbf{3.4} & 97.5 & 98.1 & 97.9 & 97.6 & \textbf{98.5} \\
051\_large\_clamp & 4.6 & \textbf{3.9} & 72.9 & 76.3 & 77.1 & 77.8 & \textbf{82.5} \\
052\_extra\_large\_clamp & 6.2 & \textbf{4.6} & 69.8 & 71.2 & 72.5 & 73.6 & \textbf{78.9} \\
061\_foam\_brick & 6.2 & \textbf{5.9} & 92.5 & 93.4 & 91.1 & 91.0 & \textbf{95.6} \\
\hline
MEAN & 6.1 & 5.3 & 93.0 & 93.6 & 93.7 & 94.1 & 95.9 \\
\hline
\end{tabular}
\end{center}
\end{table*}

\begin{table*}[h]
\caption{Comparison of surface reconstruction error and pose estimation accuracy results on on the warehouse objects.}
\label{tab:results_warehouse}
\begin{center}
\begin{tabular}{|l|c|c|c|c|c|c|c|}
\hline
& \multicolumn{2}{c|}{Reconstruction (mm)} & \multicolumn{5}{c|}{6D Pose Estimation} \\
\hline
& ElasticFusion & Object-RPE & DenseFusion (DF) & DF-PM & DF-PM-PD & DF-PM-PD-PC & Object-RPE \\
\hline
001\_frasvaf\_box & 8.3 & \textbf{6.2} & 60.5 & 63.2 & 64.1 & 65.4 & \textbf{68.7} \\
002\_small\_jacky box & 7.4 & \textbf{6.9} & 61.3 & 66.3 & 65.1 & 66.2 & \textbf{69.8} \\
003\_jacky\_box & 6.6 & \textbf{5.8} & 59.4 & 65.4 & 66.5 & 68.3 & \textbf{73.2} \\
004\_skansk\_can & 7.9 & \textbf{7.7} & 63.4 & 66.7 & 67.5 & 67.8 & \textbf{68.3} \\
005\_sotstark\_can & 7.3 & \textbf{5.9} & 58.6 & 62.4 & 65.3 & 66.2 & \textbf{69.5} \\
006\_onos\_can & 8.1 & \textbf{6.9} & 60.1 & 63.4 & 65.7 & 66.1 & \textbf{70.4} \\
007\_risi\_frutti\_box & 5.3 & \textbf{4.2} & 59.7 & 64.1 & 63.2 & 63.5 & \textbf{67.7} \\
008\_pauluns\_box & 5.8 & \textbf{5.3} & 58.6 & 62.4 & 65.9 & 66.6 & \textbf{70.2} \\
009\_tomatpure & 7.4 & \textbf{6.2} & 63.1 & 65.6 & 66.3 & 67.3 & \textbf{73.1} \\
010\_pallet & 11.7 & \textbf{10.5} & 62.3 & 64.5 & 64.6 & 66.3 & \textbf{67.4} \\
011\_half\_pallet & 12.5 & \textbf{11.4} & 58.9 & 64.1 & 63.1 & 63.4 &  \textbf{68.5} \\
\hline
MEAN & 8.0 & 7.0 & 60.5 & 64.4 & 65.3 & 66.1 & 69.7 \\
\hline
\end{tabular}
\end{center}
\end{table*}

In order to evaluate surface reconstruction quality, we compare the reconstructed model of each object to its ground truth 3D model. For every object present in the scene, we first register the reconstructed model M to the ground truth model G by a user
interface that utilizes human input to assist traditional registration techniques \cite{marion2018label}. Next, we project every vertex from M onto G and compute the distance between the original vertex and its projection. Finally, we calculate and report the mean distance $\mu_d$ over all model points and all objects.

The results of this evaluation on the reconstruction datasets are summarised in Table~\ref{tab:results-ycb} and Table~\ref{tab:results_warehouse}. Qualitative results are shown in Fig.~\ref{fig:semantic_maps}. We can see that our reconstruction system significantly outperforms the baseline. While ElasticFusion results in the lowest reconstruction errors on two YCB objects (006\_mustard\_bottle and 011\_banana\_can), our approach achieves the best performance on the remaining objects. The results show that our reconstruction method has a clear advantage of using the proposed registration cost function. In addition, we are able to keep all surfels on object instances always \textit{active}, while ElasticFusion has to segment these surfels into \textit{inactive} areas if they have not been observed for a period of time $\partial_{t}$. This means that the object surfels are updated all the time. As a result, the developed system is able to produce a highly accurate object-oriented semantic map.

\subsection{Pose Estimation Results}
\label{sec:pose_estimation_result}

We use the average closest point distance (ADD-S) metric \cite{xiang2017posecnn, wang2019densefusion} for evaluation. We
report the area under the ADD-S curve (AUC) following PoseCNN \cite{xiang2017posecnn} and DenseFusion \cite{wang2019densefusion}. The maximum threshold is set to 10cm. The object pose predicted from our system at time t is a rigid transformation from the object coordinate system $\mathcal{O}$ to the global coordinate system $\mathcal{G}$. To compare with the performance of DenseFusion, we transform the object pose to the camera coordinate system using the transformation matrix estimated from the camera tracking stage. Table~\ref{tab:results-ycb} and Table~\ref{tab:results_warehouse} present a detailed evaluation for all the 21 objects in the YCB-Video dataset and 11 objects in the warehouse dataset. Object-RPE with the full use of projected mask, depth and color images from the semantic 3D map achieves superior performance compared to the baseline single frame predictions. We observe that in all cases combining information from multiple views improved the accuracy of the pose estimation over the original DensFusion. We see an improvement of 2.3\% over the baseline single frame method with Object-RPE, from 93.6\% to 95.9\% for the YCB-Video dataset. We also observe a marked improvement, from 60.5\% for a single frame to 69.7\% with Object-RPE on the warehouse object dataset. Furthermore, we ran a number of ablations to analyze Object-RPE including (i) DenseFusion using projected masks (DF-PM) (ii) DenseFusion using projected masks and projected depth (DF-PM-PD) (iii) DenseFusion using projected masks, projected depth, and projected RGB image (DF-PM-PD-PC). DF-PM performed better than DenseFusion on both datasets (+0.6\% and +3.9\%). The performance benefit of DF-PM-PD was less clear as it resulted in a very small improvement of +0.1\% and +0.9\% over DF-PM. For DF-PM-PD-PC, performance improved additionally with +0.5\% on the YCB-Video dataset and +1.7\% on the warehouse object dataset. The remaining improvement is due to the fusion of estimates in the EKF. In regard to run-time performance, our current system does not run in real time because of heavy computation in instance segmentation, with an average computational cost of 500ms per frame.

\section{CONCLUSIONS}

We have presented and validated a mapping system that yields high quality object-oriented semantic reconstruction while simultaneously recovering 6D poses of object instances. The main contribution of this paper is to show that taking advantage of deep learning-based techniques and our semantic mapping system we are able to improve the performance of object pose estimation as compared to single view-based methods. Through various evaluations, we demonstrate that Object-RPE benefits from the use of accurate masks generated by the semantic mapping system and from combining multiple predictions based on Kalman filter. An interesting future work is to reduce the runtime requirements of the proposed system and to deal with moving objects.
\bibliographystyle{IEEEtran}
\bibliography{References}

\end{document}